\documentclass{article}
\usepackage{spconf,amsmath,epsfig}

\usepackage{setspace}
\usepackage{epstopdf}
\usepackage{amsmath,amsfonts,amssymb}
\usepackage{graphicx}
\usepackage[colorlinks=true, allcolors=blue]{hyperref}
\usepackage{tikz}
\usepackage{verbatim}
\usepackage{siunitx}
\usepackage{graphics}
\usepackage{subfig}
\usepackage{multirow}
\usepackage{multicol}
\usepackage{xspace}
\usepackage{colortbl}
\usepackage[acronym]{glossaries}
\usepackage{tabularx}
\usepackage{colortbl}

\title{Impact assessment of missing data in model predictions for Earth observation applications}
\name{
\begin{tabular}{@{}c@{}}
Francisco Mena$^{1, 2,}$\sthanks{F.Mena acknowledges support through a scholarship (RPTU). The code is at \url{https://github.com/fmenat/missingviews-study-EO}.} \quad Diego Arenas$^{2}$ \quad  Marcela Charfuelan$^{2}$ \quad Marlon Nuske$^{2}$ \quad Andreas Dengel$^{1, 2}$
\end{tabular}
}
\address{$^{1,}$University of Kaiserslautern-Landau (RPTU), Kaiserslautern, Germany\\
     $^{2}$German Research Center for Artificial Intelligence (DFKI), Kaiserslautern, Germany\\
     }

\newcolumntype{L}{>{\raggedright\arraybackslash}X}
\newcolumntype{C}{>{\centering\arraybackslash}X}
\newcolumntype{R}{>{\raggedleft\arraybackslash}X}

\newacronym{s2}{S2}{Sentinel-2}
\newacronym{s1}{S1}{Sentinel-1}
\newacronym{scl}{SCL}{scene classification layer}
\newacronym{aa}{AA}{average accuracy}
\newacronym{mvl}{MVL}{multi-view learning}
\newacronym{prs}{PRS}{performance robustness score}
\newacronym{lfmc}{LFMC}{live fuel moisture content}
\newacronym{publicyield}{CYP}{crop yield prediction}
\newacronym{cropharvestB}{CH-B}{cropharvest binary}
\newacronym{cropharvestM}{CH-M}{cropharvest multi-crop}
\newacronym{mlp}{MLP}{multi-layer perceptron}
\newacronym{eo}{EO}{Earth observation}
\newacronym{r2}{$R^2$}{coefficient of determination}

\newcommand{\highest}[1]{{\textcolor{blue}{$\mathbf{{#1}}$}}}
\newcommand{\secondhighest}[1]{{\textcolor{black}{$\mathbf{{#1}}$}}}

\begin{document}
%
\maketitle
\begin{abstract} 
Earth observation (EO) applications involving complex and heterogeneous data sources are commonly approached with machine learning models.
However, there is a common assumption that data sources will be persistently available. 
Different situations could affect the availability of EO sources, like noise, clouds, or satellite mission failures.
In this work, we assess the impact of missing temporal and static EO sources in trained models across four datasets with classification and regression tasks.
We compare the predictive quality of different methods and find that some are naturally more robust to missing data. 
The Ensemble strategy, in particular, achieves a prediction robustness up to 100\%. 
We evidence that missing scenarios are significantly more challenging in regression than classification tasks. 
Finally, we find that the optical view is the most critical view when it is missing individually.
\end{abstract}

\vspace{0.2cm}
\textit{Copyright 2024 IEEE. Published in the 2024 IEEE International Geoscience and Remote Sensing Symposium (IGARSS 2024), scheduled for 7 - 12 July, 2024 in Athens, Greece. Personal use of this material is permitted. However, permission to reprint/republish this material for advertising or promotional purposes or for creating new collective works for resale or redistribution to servers or lists, or to reuse any copyrighted component of this work in other works, must be obtained from the IEEE. Contact: Manager, Copyrights and Permissions / IEEE Service Center / 445 Hoes Lane / P.O. Box 1331 / Piscataway, NJ 08855-1331, USA. Telephone: + Intl. 908-562-3966.
}

\section{INTRODUCTION} \label{sec:intro}  

Many data-driven solutions in \gls{eo} leverage data from multiple data sources \cite{garnot2022multi,mena2022common}.
The objective of using multiple sources is to corroborate and complement the information on individual observations (or views) for the particular task. 
The literature provides evidence that the inclusion of additional data is crucial to enrich the modeling and improve the predictive quality \cite{garnot2022multi,hong2020more,mena2023comparative}.
However, the assumption that data sources are persistently available may not hold.

There are different situations in which data sources may not be available.
Specific remote sensors for \gls{eo} have a finite lifetime (e.g. based on the fuel usage), and may be affected by noise \cite{hong2020more} or clouds in the case of optical sensors \cite{garnot2022multi}. 
Besides, unexpected errors can terminate the operation earlier, such as the failure of the Sentinel-1B satellite in 2021.

Despite the research focus on more complex \gls{mvl} models \cite{mena2022common}, few works have explored the challenge of missing views, specifically missing data sources.
Srivastava et al. \cite{srivastava2019understanding} proposed a technique to retrieve a similar sample when one view is missing.
Hong et al. \cite{hong2020more} showed that the predictions of specific \gls{mvl} models worsen less when views are missing.
Gawlikowski et al. \cite{gawlikowski2023handling} showed that a missing optical view affects the predictions more than a missing radar view.
Unlike recent works, we present a study on multiple datasets involving time series and static data.

We explore the following research question: {what is the impact of missing views in \gls{mvl} models with time series and static EO sources?}
Our analysis is a special case of \textit{domain shift}, where the change is the missing data during inference.
Based on the results of prediction robustness to missing data, it allows us to formulate advice on model selection based on the task type and input views.
Furthermore, this study can serve as a way to monitor and understand the impact of specific views in trained models, as well as the sensitivity of these models to missing data.

\begin{table}[!t]
\centering
\caption{Datasets description. With ``BC'' and ``MCC'' the binary and multiclass classification, and ``Reg'' regression tasks. The last column is the spatial resolution of the target pixel.} \label{tab:data}
\small
\begin{tabularx}{\linewidth}{llrrlr} \hline
    Dataset & Task & Samples & Years & Where & Pixel \\ \hline
    \acrshort{cropharvestB} \cite{tseng2021crop} & BC &  69000  & 2016-2022 & Global & 10 m\\
    \acrshort{cropharvestM} \cite{tseng2021crop} & MCC & 29642 & 2016-2022 &  Global & 10 m \\
    \acrshort{lfmc} \cite{rao2020sar} & Reg & 2578 & 2015-2019 & USA  & 250 m\\
    \acrshort{publicyield}  \cite{perich2023pixel} & Reg & 54098 & 2017-2021 & Swiss & 10 m \\
    \hline
\end{tabularx}
\end{table}

\section{MULTI-VIEW LEARNING and Missing Views} \label{sec:methods}

\begin{table*}[!h]
\centering
\caption{Predictive quality (\acrshort{aa}) of \gls{mvl} models for different missing views scenarios and methods in the \acrshort{cropharvestB} data. Columns to the right have more missing views. We highlight the \highest{best} and \secondhighest{second} \secondhighest{best} value. }\label{tab:missing:aa:cropB}
\small
\begin{tabularx}{\linewidth}{ll|
CCCccccc} \hline
Method &  Technique & No Miss & Radar & Optical & Weather+Static & Radar+Weather+Static & Optical+Weather+Static \\ \hline
Input-concat & Impute & $0.847$ & \secondhighest{0.831} & $0.717$ & $0.674$ & $0.642$ & $0.554$  \\ 
\hline
Feature-concat & Impute & \highest{0.849} & $0.829$ & $0.730$ & $0.712$ & $0.691$ & $0.594$ \\
Feature-cca & Exemplar &  $0.829$ & $0.491$ & $0.724$ & $0.608$ & $0.543$ & $0.570$ \\
Feature-avg   & Ignore   &  \secondhighest{0.848} & \highest{0.836} & \highest{0.797} & $0.768$ & $0.729$ & $0.668$ \\
Feature-gated & Ignore   &  \highest{0.849} & \highest{0.836} & \highest{0.797} & \secondhighest{0.773} & \secondhighest{0.748} & \secondhighest{0.700} \\
\hline
Ensemble-avg & Ignore &   $0.828$ & $0.822$ & \secondhighest{0.792} & \highest{0.822} & \highest{0.824} & \highest{0.740}  \\
\hline
\end{tabularx}
\end{table*}
\begin{table*}[!h]
\centering
\caption{Predictive quality (\acrshort{aa}) of \gls{mvl} models for different missing views scenarios and methods in the \acrshort{cropharvestM} data. Columns to the right have more missing views. We highlight the \highest{best} and \secondhighest{second} \secondhighest{best} value.}\label{tab:missing:aa:cropM}
\small
\begin{tabularx}{\linewidth}{ll|
CCCccccc} \hline
Method &  Technique & No Miss & Radar & Optical & Weather+Static & Radar+Weather+Static & Optical+Weather+Static \\ \hline
Input-concat & Impute & \highest{0.738} & $0.641$ & $0.296$ & $0.534$ & \secondhighest{0.534} & $0.142$  \\ 
\hline
Feature-concat & Impute & $0.727$ & $0.624$ & $0.290$ & $0.558$ & $0.390$ & $0.159$ \\
Feature-cca & Exemplar &  $0.727$ &  $0.285$ & $0.384$ & $0.094$ & $0.107$ & $0.100$ \\
Feature-avg   & Ignore  & $0.726$ & \secondhighest{0.674} & $0.542$ & \secondhighest{0.582} & $0.529$ & \secondhighest{0.306} \\
Feature-gated & Ignore  & \secondhighest{0.734} & $0.652$ & \secondhighest{0.561} & $0.511$ & $0.440$ & \secondhighest{0.306} \\
\hline
Ensemble-avg & Ignore &  $0.715$ & \highest{0.708} & \highest{0.613} & \highest{0.711} & \highest{0.715} & \highest{0.523}   \\
\hline
\end{tabularx}
\end{table*}

A \gls{mvl} setting consists of having multiple \textit{views} as input data to a machine learning model to improve predictive quality \cite{mena2022common}. 
A view can be any set of features or data sources expressing a different perspective of each sample, such as optical or radar images, vegetation indices, terrain information or metadata.

Several works in the literature have explored various \gls{mvl} models with neural networks to achieve an optimal fusion of the data \cite{garnot2022multi,mena2023comparative}. 
Some standard \gls{mvl} models use \textit{Input}, \textit{Feature}, or \textit{Decision} fusion strategies, where the name already suggests in which part of the model architecture the fusion is placed (first, middle, or last layer respectively).
Additionally, in the \textit{Ensemble} strategy \cite{mena2023comparative}, the predictions from view-dedicated models (previously trained) are aggregated.

During inference, the occurrence of missing views can be seen as a special case of domain shift \cite{gawlikowski2023handling}. 
By lacking views, the input data deviates from the training distribution, leading to a scenario for which the model is unprepared.
However, there are some techniques applied to trained models that can mitigate this effect, which we describe below.

\noindent \textbf{Impute}. A straightforward technique is to fill in the missing views \cite{hong2020more}. Since we consider the case where the missing view is an entire data source, there are no available features for interpolation or in-painting techniques. 
Therefore, we use the average of each view in the training data as the imputation value, as it brings more information than an arbitrary value. 

\noindent \textbf{Exemplar}. Based on information retrieval, the missing view can be replaced with a similar sample via a training-set lookup. We consider the technique in \cite{srivastava2019understanding} that searches for the missing view using the available views in a shared space. 
The space is obtained with a linear projection (via CCA) from features learned by any separated \gls{mvl} model.

\noindent \textbf{Ignore}. Some \gls{mvl} models are adapted to missing views by a dynamic fusion. 
In the Ensemble strategy, the predictions of the view-dedicated model associated to the missing view are omitted in the aggregation.
Similarly, with Feature fusion, the features of the missing views are ignored when using average as the merge function. 
Further, we include a \gls{mvl} model with gated fusion \cite{mena2023comparative} that re-normalizes the gated weights based on the available views (it omits the missing ones).

\section{EVALUATION} \label{sec:evaluation}

\subsection{Datasets} \label{sec:evaluation:data}

In the following we describe the four datasets used in this study, the Table~\ref{tab:data} presents some characteristics of these.

\noindent \textbf{\Gls{cropharvestB}}: We use the cropharvest data for multi-view crop recognition \cite{tseng2021crop}. 
This is a binary task in which the presence or absence of a specific crop at a given location during a particular season is predicted.
The input views are optical (from S2), radar (from S1), and weather time series. 
These temporal views were re-sampled monthly for 1 year.
An additional static view is the topographic information. 

\noindent \textbf{\Gls{cropharvestM}}: We use a multi-crop version of the cropharvest data with 10 classes, see \cite{tseng2021crop} for details. 
We use the same input views as for the \gls{cropharvestB} data.

\noindent \textbf{\Gls{lfmc}}: We use a dataset for multi-view moisture content estimation \cite{rao2020sar}. This is a regression task in which the vegetation water per dry biomass (in percentage) in a given location at a specific moment is predicted.
The input views are optical (from L8) and radar (from S1) time series. 
These views were re-sampled monthly along 4 months. 
Additional static views are the topographic information, soil features, canopy height, and land-cover class. 

\noindent \textbf{\Gls{publicyield}}: We use a dataset for multi-view crop yield estimation \cite{perich2023pixel}, focused on cereals. This is a regression task in which the amount of crop (in t/ha) grown in a particular location during a growing season is predicted.
The input views are optical (from S2), and weather time series.
The weather view was aligned to the optical time series (5-days sample) from seeding to harvesting (variable length).

\begin{table*}[!t]
\centering
\caption{Predictive quality (\acrshort{r2}) of \gls{mvl} models for different missing views scenarios and methods in the \acrshort{lfmc} data. Columns to the right have more missing views. $\dagger$ is a value lower than $-100$. We highlight the \highest{best} and \secondhighest{second} \secondhighest{best} value.}\label{tab:missing:r2:lfmc}
\small
\begin{tabularx}{\linewidth}{ll|CCCCcc} \hline
Method &  Technique & No Miss & Radar & Optical & Statics & Radar+Statics & Optical+Statics \\ \hline
Input-concat & Impute & \secondhighest{0.717} & \secondhighest{0.650} & $0.060$ & $0.185$ & $0.165$ & $-0.047$ \\ 
\hline
Feature-concat & Impute & $0.667$ & $0.599$ & \highest{0.274} & \secondhighest{0.352} & \secondhighest{0.290} & \secondhighest{0.081}  \\
Feature-cca & Exemplar &  $0.667$ &  $\dagger$ & $-0.260$ & $\dagger$ & $\dagger$ & $\dagger$ \\
Feature-avg   & Ignore  & $0.683$ & $0.618$ & $0.142$ &  $\dagger$   &  $\dagger$ & $\dagger$ \\
Feature-gated & Ignore  & \highest{0.737} & \highest{0.651} & $0.138$ & $-0.422$ & $-5.326$ & $-1.693$ \\
\hline
Ensemble-avg & Ignore &  $0.312$ & $0.292$ & \secondhighest{0.243} & \highest{0.407} & \highest{0.392} & \highest{0.239}  \\
\hline
\end{tabularx}
\end{table*}

\begin{table}[!t]
\centering
\caption{Predictive quality (\acrshort{r2}) of \gls{mvl} models for different missing views scenarios and methods in the \acrshort{publicyield} data. With $\dagger < -100$. We highlight the \highest{best} and \secondhighest{second} \secondhighest{best} value.}\label{tab:missing:r2:yield}
\small
\begin{tabularx}{\linewidth}{ll|cCC} \hline
Method &  Technique & No Miss & Optical & Weather \\ \hline
Input-concat & Impute & $0.823$ & $-1.435$ &  $0.710$  \\ 
\hline
Feature-concat & Impute & \secondhighest{0.827} & $-1.058$ & \secondhighest{0.760} \\
Feature-cca & Exemplar &  \secondhighest{0.827} &  $-0.343$ & $\dagger$ \\ 
Feature-avg   & Ignore   & \highest{0.828} & $-0.993$ & $-8.664$ \\
Feature-gated & Ignore   &  $0.823$ & \secondhighest{0.193} & $-7.534$ \\
\hline
Ensemble-avg & Ignore & $0.768$ & \highest{0.593} &  \highest{0.823}  \\
\hline
\end{tabularx}
\end{table}

\subsection{Experiment Settings} \label{sec:evaluation:conf}

We apply a z-score normalization to the input data.
The categorical and ordinal views (land-cover and canopy height) are one-hot-vector encoded. 
We use a \gls{mlp}, as the encoder for the static views.
For the temporal views, we use a 1D convolutional network, except in the \gls{lfmc} data, where we use a recurrent network (with gated recurrent units) as an encoder. 
We use two layers with 128 dimensions in the encoders, and a \gls{mlp} with one hidden layer of 128 units as prediction head. 
An ADAM optimizer is used with batch-size 128 
and early stopping. %
The loss function is cross-entropy in classification and mean squared error in regression task. 

We train and evaluate using the 10-fold cross-validation. 
The predictive quality is measured with the \gls{aa} in classification, and the \gls{r2} in regression. 
We include the \gls{prs} presented by Heinrich et al. \cite{heinrich2023targeted}. 
It is based on the error in the predictions with missing views relative to the same error when all views are available.

\subsection{Missing View Scenarios} \label{sec:evaluation:miss}

The assessment is based on models trained with all the views. 
The missing views scenario consists of making predictions on the validation fold with fewer views available than during training.
We experiment with a moderate degree of missingness, as the case when only radar is missing, and when optical is missing; An intermediate missingness, when all other views except radar and optical are missing (inference with radar and optical); And an extreme degree of missingness, when all views missing except one: a single-view inference with only radar or optical. 
We compare the techniques described in Sec.~\ref{sec:methods}.
Two \gls{mvl} models with the Impute technique: Input and Feature with concatenation (Input-concat, Feature-concat).
Three \gls{mvl} models with Ignore techniques: Feature and Ensemble with averaging (Feature-avg, Ensemble-avg), and Feature with gated fusion (Feature-gated).
Lastly, one \gls{mvl} model based on the Feature fusion with the Exemplar technique (Feature-cca, see Sec.~\ref{sec:methods} for details).

\subsection{Experiment Results} \label{sec:evaluation:results}

In Tables~\ref{tab:missing:aa:cropB}-\ref{tab:missing:aa:cropM} we show the predictive quality in classification tasks.
The results of the Input-concat method decreases significantly when views are missing.
We observe that, when using Feature fusion with Ignore techniques (Feature-avg, Feature-gated) the impact of missing views is mitigated more than with Impute or Exemplar (Feature-concat, Feature-cca). 
However, these do not achieve the values of the Ensemble-avg, which is the method least affected by missing views. 

We observe similar results in the regression tasks, shown in Tables~\ref{tab:missing:r2:lfmc}-\ref{tab:missing:r2:yield}, except when using the Ignore techniques. 
The predictions of Feature fusion-based models with Ignore techniques become worse, up to negative \gls{r2} values, when views are missing.
Besides, the results of the Ensemble-avg method in the \gls{lfmc} data is relatively bad ($ \approx 0.3$) with no or a moderate degree of missing views.

\begin{figure}[!t]
    \centering
    \includegraphics[width=\linewidth]{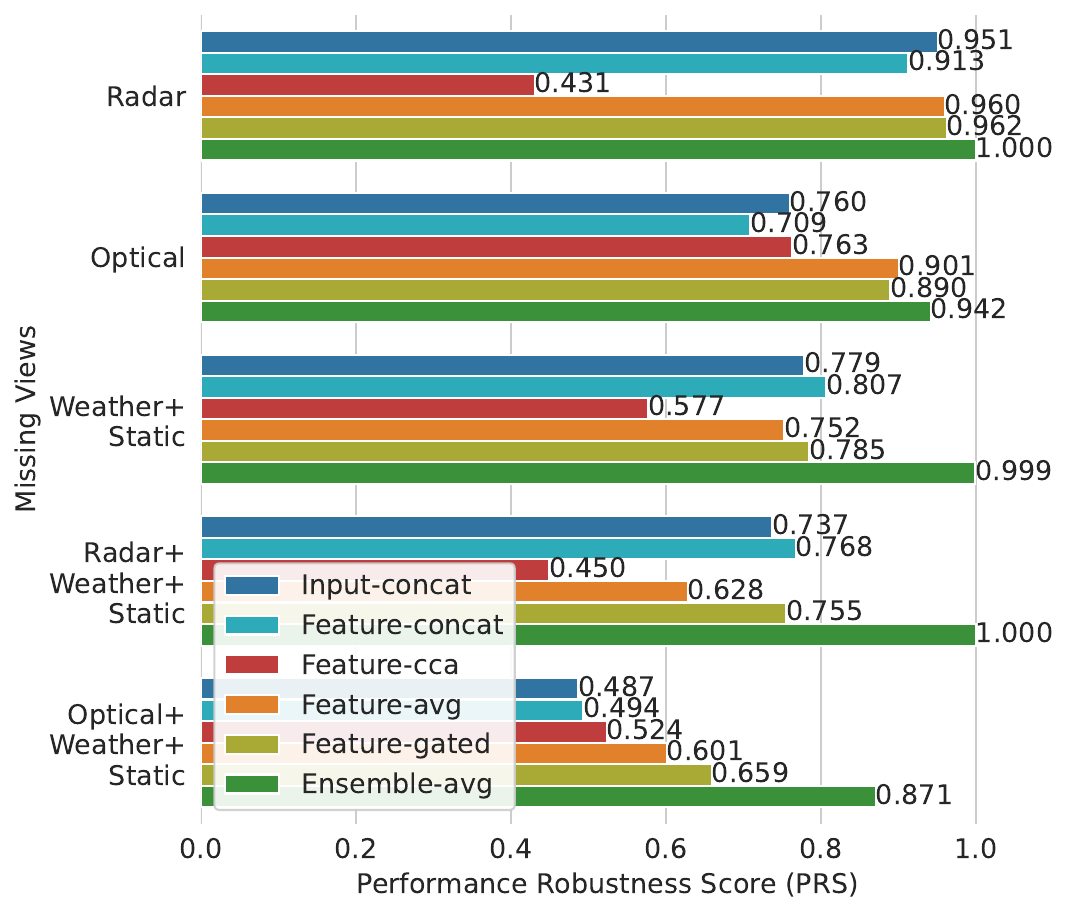}
    \caption{Prediction robustness of \gls{mvl} models for different missing views scenarios in the \gls{cropharvestB} data (classification).}
    \label{fig:prs:cropbinary}
\end{figure}
\begin{figure}[!t]
    \centering
    \includegraphics[width=\linewidth]{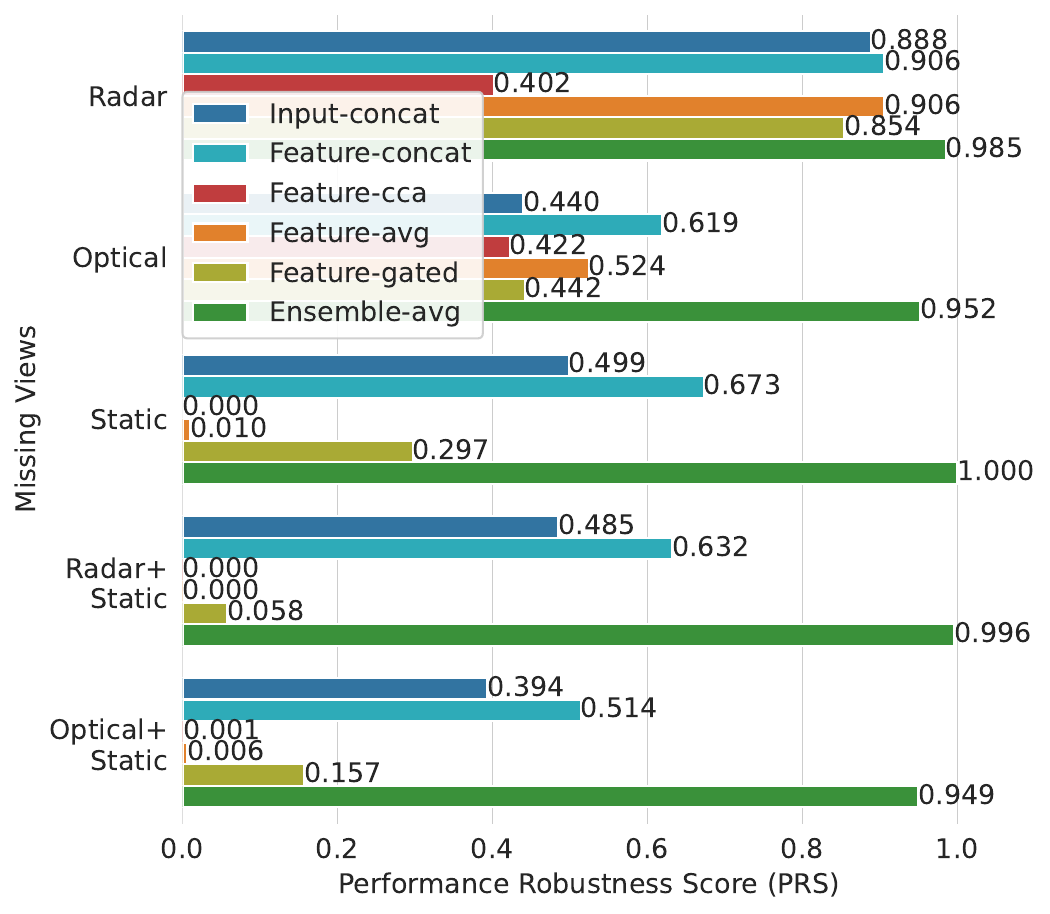}
    \caption{Prediction robustness of \gls{mvl} models for different missing views scenarios in the \gls{lfmc} data (regression).}
    \label{fig:prs:lfmc}
\end{figure}

For the robustness results in Fig.~\ref{fig:prs:cropbinary} and \ref{fig:prs:lfmc}, we corroborate the lower impact of missing views  in the \gls{mvl} models when using the Ignore techniques.
The Ensemble-avg method got a \gls{prs} close to one in some cases, which means that the error of predictions with missing views is lower or same than the error in predictions without missing views.
However, this is on average, as there can still be negative changes in the predictive quality, such as when radar is missing in the \gls{cropharvestB} data (Table~\ref{tab:missing:aa:cropB}, with PRS of one), or relatively bad predictive quality, such as in the \gls{lfmc} data (Table~\ref{tab:missing:r2:lfmc}).
We notice that the Feature-concat method has higher robustness than Feature-avg and Feature-gated in regression.
Besides, the Feature-cca method has a fairly low robustness, especially in regression tasks, reaching a PRS of 0 in some scenarios.

Overall, we note that the impact of missing views depends on the \gls{mvl} model along with how to treat missing views, as previous works have shown \cite{hong2020more,garnot2022multi,gawlikowski2023handling}.
The negative effect of missing view increases from the moderate, intermediate to the extreme degree. 
In addition, we observe that the impact of missing optical view is stronger than when missing radar.
This means that the optical view is more difficult to supplement than others, reflecting its greater importance for the EO applications studied.
Ancillary data, like static and weather views, still provide valuable information to \gls{mvl} models. For instance, the predictive quality of some methods in the \gls{lfmc} data become quite worse when the static views are missing, and, in \gls{cropharvestB}, some become much worse when weather and static views are missing compared to the optical view.

\section{CONCLUSION} \label{sec:conclu}

In this work, we evaluated the impact of missing views in \gls{mvl} models across various tasks with time series and static \gls{eo} data.
We showed that missing specific views (such as optical) significantly affects the predictive quality, 
whereas, as the number of missing views increases, the negative effect does as well.
Nevertheless, the prediction robustness can be improved by designing a method adjustable for the missing views. 
In addition, due to the differences in predicting a continuous value to a categorical one, the impact of missing views is more severe in regression than in classification tasks.
Based on the results, we provide the following advice for model selection in missing view scenarios:
if views are sufficiently discriminative to allow individual predictions of the task, use the Ensemble strategy that ignores the missing predictions, otherwise use the Feature fusion strategy ignoring missing views in classification, or imputing missing views in regression tasks.
Since we explored the case of trained \gls{mvl} models, future research would focus on modifying model learning.

\bibliographystyle{IEEEbib}
\bibliography{main}

\begin{thebibliography}{10}

\bibitem{garnot2022multi}
V.~S.~F. Garnot, L.~Landrieu, and N.~Chehata,
\newblock ``Multi-modal temporal attention models for crop mapping from
  satellite time series,''
\newblock {\em ISPRS Journal of Photogrammetry and Remote Sensing}, vol. 187,
  pp. 294--305, 2022.

\bibitem{mena2022common}
F.~Mena, D.~Arenas, M.~Nuske, and A.~Dengel,
\newblock ``Common practices and taxonomy in deep multi-view fusion for remote
  sensing applications,''
\newblock {\em IEEE Journal of Selected Topics in Applied Earth Observations
  and Remote Sensing}, pp. 1--21, 2024.

\bibitem{hong2020more}
D.~Hong, L.~Gao, N.~Yokoya, J.~Yao, J.~Chanussot, Q.~Du, and B.~Zhang,
\newblock ``More diverse means better: {Multimodal} deep learning meets
  remote-sensing imagery classification,''
\newblock {\em IEEE Transactions on Geoscience and Remote Sensing}, vol. 59,
  no. 5, pp. 4340--4354, 2020.

\bibitem{mena2023comparative}
F.~Mena, D.~Arenas, M.~Nuske, and A.~Dengel,
\newblock ``A comparative assessment of multi-view fusion learning for crop
  classification,''
\newblock in {\em IEEE International Geoscience and Remote Sensing Symposium
  (IGARSS)}. IEEE, 2023, pp. 5631--5634.

\bibitem{srivastava2019understanding}
S.~Srivastava, J.~E. Vargas-Munoz, and D.~Tuia,
\newblock ``Understanding urban landuse from the above and ground perspectives:
  {A} deep learning, multimodal solution,''
\newblock {\em Remote Sensing of Environment}, vol. 228, pp. 129--143, 2019.

\bibitem{gawlikowski2023handling}
J.~Gawlikowski, S.~Saha, J.~Niebling, and X.~X. Zhu,
\newblock ``Handling unexpected inputs: {Incorporating} source-wise
  out-of-distribution detection into {SAR}-optical data fusion for scene
  classification,''
\newblock {\em EURASIP Journal on Advances in Signal Processing}, vol. 2023,
  no. 1, pp. 47, 2023.

\bibitem{tseng2021crop}
G.~Tseng, I.~Zvonkov, C.~L. Nakalembe, and H.~Kerner,
\newblock ``{{CropHarvest}}: {{A}} global dataset for crop-type
  classification,''
\newblock {\em Proceedings of NIPS Datasets and Benchmarks Track}, 2021.

\bibitem{rao2020sar}
K.~Rao, A.~P. Williams, J.~F. Flefil, and A.~G. Konings,
\newblock ``{SAR}-enhanced mapping of live fuel moisture content,''
\newblock {\em Remote Sensing of Environment}, vol. 245, pp. 111797, 2020.

\bibitem{perich2023pixel}
G.~Perich, M.~O. Turkoglu, L.~V. Graf, J.~D. Wegner, H.~Aasen, A.~Walter, and
  F.~Liebisch,
\newblock ``Pixel-based yield mapping and prediction from {Sentinel-2} using
  spectral indices and neural networks,''
\newblock {\em Field Crops Research}, vol. 292, pp. 108824, 2023.

\bibitem{heinrich2023targeted}
R.~Heinrich, C.~Scholz, S.~Vogt, and M.~Lehna,
\newblock ``Targeted adversarial attacks on wind power forecasts,''
\newblock {\em Machine Learning}, 2023.

\end{thebibliography}

\end{document}